\documentclass{article}

\usepackage[preprint]{neurips_2024}

\usepackage[utf8]{inputenc} % allow utf-8 input
\usepackage[T1]{fontenc} % use 8-bit T1 fonts
\usepackage{hyperref} % hyperlinks
\usepackage{url} % simple URL typesetting
\usepackage{booktabs} % professional-quality tables
\usepackage{amsfonts} % blackboard math symbols
\usepackage{nicefrac} % compact symbols for 1/2, etc.
\usepackage{microtype} % microtypography
\usepackage{xcolor} % colors
\usepackage{amsmath}
\usepackage{graphicx}
\usepackage{caption}
\usepackage{wrapfig, lipsum}
\usepackage{multirow}
\usepackage{listings}
\usepackage{subcaption}
\usepackage{longtable}
\usepackage{natbib}
\setcitestyle{numbers,square}

\definecolor{codegreen}{rgb}{0,0.6,0}
\definecolor{codegray}{rgb}{0.5,0.5,0.5}
\definecolor{codepurple}{rgb}{0.58,0,0.82}
\definecolor{backcolour}{rgb}{0.95,0.95,0.92}

\lstdefinestyle{mystyle}{
    backgroundcolor=\color{backcolour},   
    commentstyle=\color{codegreen},
    keywordstyle=\color{magenta},
    numberstyle=\tiny\color{codegray},
    stringstyle=\color{codepurple},
    basicstyle=\ttfamily\footnotesize,
    breakatwhitespace=false,         
    breaklines=true,                 
    captionpos=b,                    
    keepspaces=true,                 
    numbers=left,                    
    numbersep=5pt,                  
    showspaces=false,                
    showstringspaces=false,
    showtabs=false,                  
    tabsize=2
}

\title{MoE Parallel Folding: Heterogeneous Parallelism Mappings for Efficient Large-Scale MoE Model Training with Megatron Core}

    \author{
      \textbf{Dennis Liu}\thanks{These authors contributed equally to this work. Authors are listed in alphabetical order.}
      \quad
      \textbf{Zijie Yan}$^{*}$ 
      \quad
      \textbf{Xin Yao}
      \quad
      \textbf{Tong Liu}
      \\
      \textbf{Vijay Korthikanti} 
      \quad
      \textbf{Evan Wu}
      \quad 
      \textbf{Shiqing Fan}
      \quad 
      \textbf{Gao Deng}
      \quad
      \textbf{Hongxiao Bai}
      \\
      \textbf{Jianbin Chang}
      \quad
      \textbf{Ashwath Aithal}
      \quad
      \textbf{Michael Andersch}
      \quad
      \textbf{Mohammad Shoeybi}
       \\
      \textbf{Jiajie Yao}
      \quad
      \textbf{Chandler Zhou}
      \quad
      \textbf{David Wu}
      \quad
      \textbf{Xipeng Li}
      \quad
      \textbf{June Yang} \thanks{Corresponding author: juney@nvidia.com} \\
      \textbf{NVIDIA} \\
     \{denliu, zijiey, xiny, tongliu, vkorthikanti, evwu, shiqingf, gdeng, hongxiaob, jianbinc, aaithal, \\ mandersch, mshoeybi, jiajiey, chandlerz, davidwu, xipengl, juney\}@nvidia.com
    }
    
\begin{document}
      \maketitle
      \setcounter{footnote}{0}

      \begin{abstract}
    Mixture of Experts (MoE) models enhance neural network scalability by dynamically selecting relevant experts per input token, enabling larger model sizes while maintaining manageable computation costs. However, efficient training of large-scale MoE models across thousands of GPUs presents significant challenges due to limitations in existing parallelism strategies.
    We introduce an end-to-end training framework for large-scale MoE models that utilizes five-dimensional hybrid parallelism: Tensor Parallelism, Expert Parallelism, Context Parallelism, Data Parallelism, and Pipeline Parallelism. Central to our approach is MoE Parallel Folding, a novel strategy that decouples the parallelization of attention and MoE layers in Transformer models, allowing each layer type to adopt optimal parallel configurations.
    Additionally, we develop a flexible token-level dispatcher that supports both token-dropping and token-dropless MoE training across all five dimensions of parallelism. This dispatcher accommodates dynamic tensor shapes and coordinates different parallelism schemes for Attention and MoE layers, facilitating complex parallelism implementations.
    Our experiments demonstrate significant improvements in training efficiency and scalability. We achieve up to 49.3\% Model Flops Utilization (MFU) for the Mixtral 8x22B model and 39.0\% MFU for the Qwen2-57B-A14B model on H100 GPUs, outperforming existing methods. The framework scales efficiently up to 1,024 GPUs and maintains high performance with sequence lengths up to 128K tokens, validating its effectiveness for large-scale MoE model training.
    The code is available in Megatron-Core \footnote{https://github.com/NVIDIA/Megatron-LM}.
      \end{abstract}

  \section{Introduction}
  % Intro to MoE
  In recent years, Mixture of Experts (MoE) models have emerged as a powerful
  architecture for scaling neural networks to unprecedented sizes. By leveraging
  multiple experts in one model and dynamically selecting the most relevant experts
  for each input token, MoE models can accommodate larger parameter counts while maintaining manageable computational costs. This approach not only enhances model capacity but also improves
  performance on a variety of tasks compared to their dense counterparts. Recent
  MoE models scale to hundreds of billions or even trillions of parameters and
  achieve state-of-the-art performance~\cite{mixtral, deepseek-v2, qwen2,grok1_2024,databricks_dbrx_2024, snowflake_arctic}.

  % Distributed LLM training and limits
  Training large-scale MoE models, however, presents significant challenges. As the
  model size increases, efficient distributed training across thousands of GPUs
  becomes essential. Different parallelism strategies have been proposed in recent years for distributed
  LLM training, including model parallelism, data parallelism, and pipeline
  parallelism~\cite{Megatron-TP,zero-dp,Pipedream}. However, a single parallelism strategy has limitations regarding
  scalability. For example, the performance of data parallelism with ZeRO-3 will
  decrease dramatically when the number of GPUs increases to several thousands
  \cite{Megatron-3d}.

  % Hybrid parallelism and its challenges to MoE
  To address the growing computational and memory requirements of increasingly large models, hybrid parallelism—which integrates multiple parallelization strategies—has become essential. While 3D parallelism is widely adopted for training large-scale dense models, optimizing training efficiency for MoE models using hybrid parallelism presents greater complexity. This is primarily due to the inherent sparsity of MoE models, which results in a significantly lower computation-to-parameter ratio compared to dense models. Employing a small degree of model parallelism often leads to out-of-memory issues for MoE models, whereas a large degree introduces substantial communication overhead and diminishes computational efficiency.

  Previous approaches have predominantly relied on expert parallelism coupled with data parallelism to scale the training of MoE models. However, the diversity in the number of experts, the size of individual experts, and the sequence length of training samples across different MoE models necessitates an adaptive parallelism strategy tailored to each scenario. Moreover, the distinct computational characteristics of the Attention and Feed-Forward Network (FFN) layers in MoE models render a uniform parallelism strategy across these layers suboptimal. The dynamic nature of MoE models, including on-the-fly token routing and variable tensor shapes, further exacerbates the complexity of designing efficient training algorithms. These challenges collectively underscore the need for a coordinated integration of multiple parallelism strategies to optimize the training of MoE models.

  To address these challenges, we propose an end-to-end training framework for large-scale MoE models based on 5-D hybrid parallelism, which integrates five key parallelism dimensions: Tensor Parallelism(TP), Expert Parallelism(EP), Context Parallelism(CP), Data Parallelism(DP), and Pipeline Parallelism(PP). At the core of our framework are two innovations: MoE Parallel Folding and an efficient token-level dispatcher. MoE Parallel Folding is a novel hybrid parallelism strategy that disentangles the parallel mappings of the Attention and MoE components in Transformer-based models. Our key insight is that enabling flexible and distinct parallelism configurations for these components unlocks a comprehensive parallelism space, ensuring optimal performance. Additionally, to support arbitrary parallelism combinations while maintaining numerical correctness, we designed a highly flexible and efficient token-level MoE dispatcher. This dispatcher accommodates both token-dropping and token-dropless training paradigms, eliminates sequence length dependencies, and enables dynamic tensor shapes, thereby facilitating the implementation of complex parallelism schemes.

  The contributions of this paper are as follows:
  \begin{enumerate}
    \item MoE Parallel Folding: We introduce MoE Parallel Folding, the first approach
      that decouples parallelization strategies for attention and MoE layers,
      enabling each layer to adopt its own optimal configurations. This method enables
      the folding of communication-intensive parallel dimensions to fit within high-bandwidth
      intra-node networks, reducing communication overhead.

    \item Flexible and efficient token-level dispatcher: We develop a novel
      dispatcher that supports both token-dropping and token-dropless MoE training
      with five-dimensional hybrid parallelism, including TP, EP, CP, DP, and PP.
    % This flexibility overcomes the limitations of existing frameworks, enhancing scalability for large-scale MoE models.

    \item Performance enhancements: Through MoE Parallel Folding, we demonstrate
      significant improvements in training efficiency and scalability for large-scale
      MoE models. By optimizing the utilization of network resources based on
      model characteristics, we achieve 49.3\% MFU for Mixtral 8x22B and 39.0\% MFU
      for Qwen2-57B-A14B on H100 GPUs.
  \end{enumerate}

  \section{Related Work}
  \subsection{Mixture of Experts}
  The Mixture of Experts (MoE) architecture~\cite{Shazeer2017OutrageouslyLN}
  enhances neural network capacity and efficiency by integrating multiple
  specialized sub-networks, known as experts, under the supervision of a routing
  mechanism. Each expert specializes in different regions of the input space or captures
  distinct features of the data. The router dynamically selects the most
  relevant experts for each input, enabling the model to process a diverse range
  of patterns more effectively than traditional monolithic architectures. This
  selective activation allows MoE models to scale significantly without a
  proportional increase in computational complexity, as only a subset of experts
  is engaged per input.

  Incorporating MoE architectures into Transformer models~\cite{transformer} has
  led to substantial advancements, achieving superior performance compared to
  dense counterparts. Notable examples include GShard~\cite{gshard}, which
  scaled models to trillions of parameters using MoE layers, and the Switch Transformer~\cite{switch-transformer},
  which improved scalability and efficiency with a streamlined routing mechanism.
  Similarly, GLaM~\cite{glam} demonstrated the effectiveness of MoE in large-scale
  language modeling tasks. These models leverage the sparsity introduced by MoE to
  maintain manageable computational demands, activating only a subset of experts
  per input to reduce overall computation and memory requirements.

  Addressing challenges such as load balancing among experts and managing
  dynamic computation graphs is critical for MoE architectures. Traditional methods
  often employ token-dropping training~\cite{switch-transformer}, setting a capacity
  factor for each expert to prevent overloading. While this mitigates performance bottlenecks,
  it can result in some tokens being dropped or not fully processed, potentially
  affecting model quality. In contrast, Megablocks~\cite{megablocks} utilizes
  token-dropless training to ensure all input tokens are processed,
  demonstrating better performance for models of equivalent size and training
  data by avoiding the loss of information inherent in token dropping.

  Recent developments focus on fine-grained MoE architectures to further enhance
  performance~\cite{fine-grained-law, FineGrained}. Approaches like DeepSeek-MoE~\cite{deepseek,deepseek-v2} segment experts into
  smaller sub-experts and activate a greater number of experts per token,
  achieving higher degrees of specialization. This fine-grained specialization allows
  models to capture complex patterns and relationships within the data more effectively.

  \subsection{Distributed MoE Training}
The substantial size of Large Language Models (LLMs) often exceeds the memory and computational capacity of a single GPU, necessitating distributed training strategies to manage resource constraints effectively. Conventional distributed training methods include TP, DP, CP and PP. TP divides the computations of neural network layers across multiple devices, allowing for parallel processing of tensors within layers\cite{Megatron-TP}. TP can significantly reduce the memory consumption of each model rank but introduces some intra-layer communication overhead. DP distributes batches of data across replicas of the model on different devices, aggregating gradients during training\cite{DP}. Zero Redundancy Optimizer(ZeRO) further splits optimizer states, model weights and gradients across DP group to trade memory with communication~\cite{DP-pytorch, zero-dp}. CP splits the input sequences into small segments for each device, allowing for very long sequence length training\cite{Ulysses, ring-attention}.
PP splits\cite{PP-alibaba, Megatron-3d, Pipedream, PP-memory-efficient} the model layers across devices, enabling different stages of the model to process data concurrently in a pipelined fashion.

In the context of MoE models, EP is employed to optimize MoE training by assigning different experts to different devices\cite{glam, switch-transformer, EP-others, gshard, Shen2022MoESysAD}. During training, the routing mechanism directs inputs to the appropriate experts across devices. EP efficiently utilizes hardware resources by balancing the computational load and reducing inter-device communication overhead associated with expert data exchanges.
To further enhance the efficiency of distributed MoE training, hybrid parallelism strategies are leveraged which combines EP with other parallelism methods, like FSDP and TP\cite{TED, Deepspeed-MoE, Hwang2022TutelAM, megablocks}.

  \section{Method}
  \subsection{Preliminary}

  \subsubsection{Mixture of Experts}
The Mixture of Experts (MoE) is a neural network architecture that dynamically selects the most relevant experts to process each individual token. The MoE layer consists of $E$ expert networks and a gating network, which determines the routing by computing a selection probability for each expert. The output of the MoE layer is computed as a weighted aggregation of the outputs from the selected experts, based on their gating probabilities:
\begin{equation}
    \mathbf{y} = \sum_{e=1}^{E} g_{e}(\mathbf{x}) \cdot f_{e}(\mathbf{x}),
\end{equation}
where $g_{e}(\mathbf{x})$ denotes the gating weight for expert $e$, and $f_{e}(\mathbf{x})$ represents the output of expert $e$. Among the various gating strategies, the \textbf{Top-K gating} method is the most widely used:
\begin{equation}
    g_{e}(\mathbf{x}_i) = 
    \begin{cases}
        s_{i} & \text{if } i \in \text{TopK}(\mathbf{s}, K), \\
        0 & \text{otherwise},
    \end{cases}
\end{equation}
where $\mathbf{x}_i$ is the $i$-th token input to the current expert, and $\mathbf{s}$ is computed as follows:
\begin{equation}
    \mathbf{s} = G(W_g \cdot \mathbf{x}).
\end{equation}
Here, $W_g$ represents the weight matrix of the gating network, and $G$ denotes a non-linear activation function.

The learnable gating network enables dynamic routing, which can lead to load imbalance issues. To mitigate this, in addition to the auxiliary loss, a \textbf{capacity factor} ($CF$) is introduced to regulate load balancing among experts. The capacity factor defines the maximum capacity of each expert relative to the average expected load, ensuring that computational resources are evenly distributed and preventing bottlenecks caused by uneven workloads. The capacity per expert is calculated as:
\begin{equation}
    \text{Capacity per Expert} = CF \cdot \frac{L}{E},
\end{equation}
where $L$ is the total number of tokens to process, $E$ is the number of experts, and $CF \geq 1$ is the capacity factor. Tokens that exceed the capacity of a given expert are dropped.

  \subsubsection{Expert Parallelism}
  In large-scale distributed training of MoE models, EP
  efficiently distributes computation across multiple devices by assigning distinct experts to each device.
  This parallelization strategy reduces communication overhead
  while optimizing hardware utilization. The EP process comprises three key stages:

  \paragraph{Token Dispatching}
  Initially, input tokens are grouped according to their assigned experts through data permutation,
  ensuring tokens destined for the same expert are stored contiguous in memory.
  An \emph{All-to-All} collective communication operation then exchanges token data between
  devices, allowing each device to receive only the tokens required by its locally hosted experts.

  \paragraph{Expert Computation} 
  Each device processes its local batch of tokens through its designated experts in parallel.
  Since experts operate independently, this stage requires no inter-device communication,
  allowing for efficient concurrent computation across the distributed system.

  \paragraph{Token Restore}
  After expert processing, the output tokens are rearranged to restore their original sequence
  order through an inverse permutation operation. This restoration step ensures proper
  alignment for subsequent layer operations while maintaining the model's sequential processing
  requirements. The restored tokens can then flow into the next layer of the network.

  \begin{figure}[t]
    \centering
    \includegraphics[width=\textwidth]{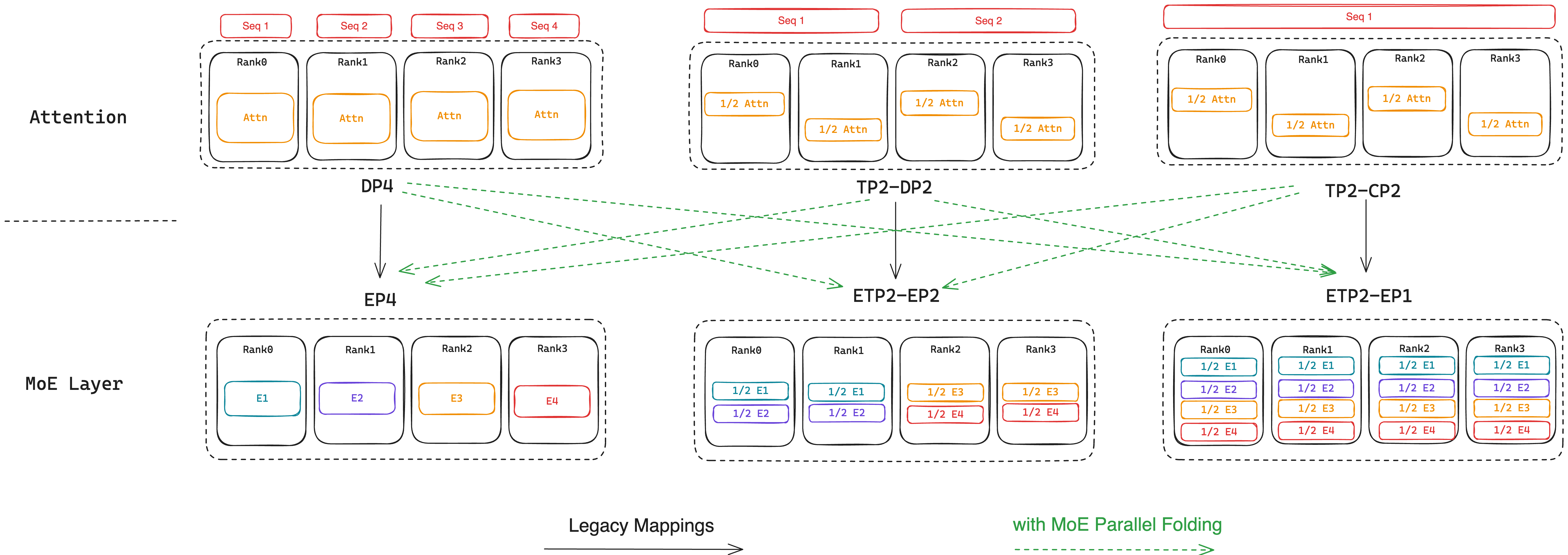}
    \caption{Illustration of parallelism mappings with MoE Parallel Folding.}
    \label{fig:parallel-folding}
  \end{figure}

  \subsection{MoE Parallel Folding}
  % Motivation
  Attention layers and MoE layers in Transformers exhibit distinct 
  computation and communication patterns. Attention operations are performed at the 
  whole-sequence level with dense computation, requiring information exchange between 
  devices holding sub-sequences when using TP and CP. In contrast, MoE layers process 
  individual tokens rather than whole sequences, and their inherent sparsity makes them 
  more suitable for EP with lower communication overhead.

  Consequently, forcing MoE layers to follow the same parallelism mapping as Attention 
  layers is sub-optimal. To achieve optimal hybrid parallelism for MoE models, 
  we propose MoE Parallel Folding, which disentangles the parallel mappings between 
  Attention and MoE layers.

  % Illustration with figures
  As shown in Figure~\ref{fig:parallel-folding}, previous methods place the EP group
  in a sub-group of DP, which greatly restricts the scalability of MoE. The maximum
  degree of expert parallelism is bounded by the degree of data parallelism.
  Instead, we flatten the parallelism mappings of the attention layer and allow model parallelism
  in the MoE layer to be folded with arbitrary sub-groups of attention, making
  the parallelism mappings of MoE layer more flexible and efficient.

  % Implementation
  Specifically, for the attention layers, we form a four-dimensional parallel
  group comprising $TP \times CP \times DP \times PP$. For the MoE layers, we
  define another four-dimensional group consisting of
  $TP \times EP \times DP \times PP$. For convenience, we name the TP and DP group for MoE layer as Expert-TP(ETP) and Expert-DP(EDP). 
  The only restriction is that the number of PP groups and members of each PP group for the Attention and MoE layer must be consistent.
  This separation allows us to set flexible and independent parallelism configurations for attention and MoE layers.

  % Benefits
  MoE Parallel Folding provides two main benefits. First, it allows selecting the optimal parallelism mapping for the MoE layer independently of the Attention layer. For example, EP
  is more communication-efficient than ETP. We can replace ETP with EP
  and fold it with TP in the Attention layer. Second, the folded parallelism mappings
  enable communication within more compact groups. By folding model parallelism across
  attention and MoE layers, the scope of intra-layer communication is reduced, allowing it to fit within high-bandwidth intra-node connections more effectively.

  \subsection{Flexible and Efficient Token Dispatcher}
  % The necessity of token-level dispatching to make MoE model support different parallelisms for Attention and MoE layer.
  Arbitrary hybrid parallelism with MoE Parallel Folding necessitates a flexible
  and scalable token dispatcher. The dispatcher is responsible for routing
  tokens to their assigned experts across various parallelism dimensions. To ensure
  numerical correctness while maintaining high performance under different parallelism
  strategies, we have designed a unified token dispatcher that handles both ETP and EP within the MoE layer.

  % Workflow of token-level dispatching.
  With MoE Parallel Folding, the inputs fed into the MoE layer from the
  attention layer are split either along the batch dimension (DP)
  or the sequence dimensions (CP and TP). In both
  scenarios, different ranks contain different chunks of tokens. Since the expert
  layer computes the features of each token individually, we can employ the same
  workflow for the token dispatcher regardless of the parallelism mappings of
  the attention layer.

\begin{figure}[t]
    \centering
    \includegraphics[width=\textwidth]{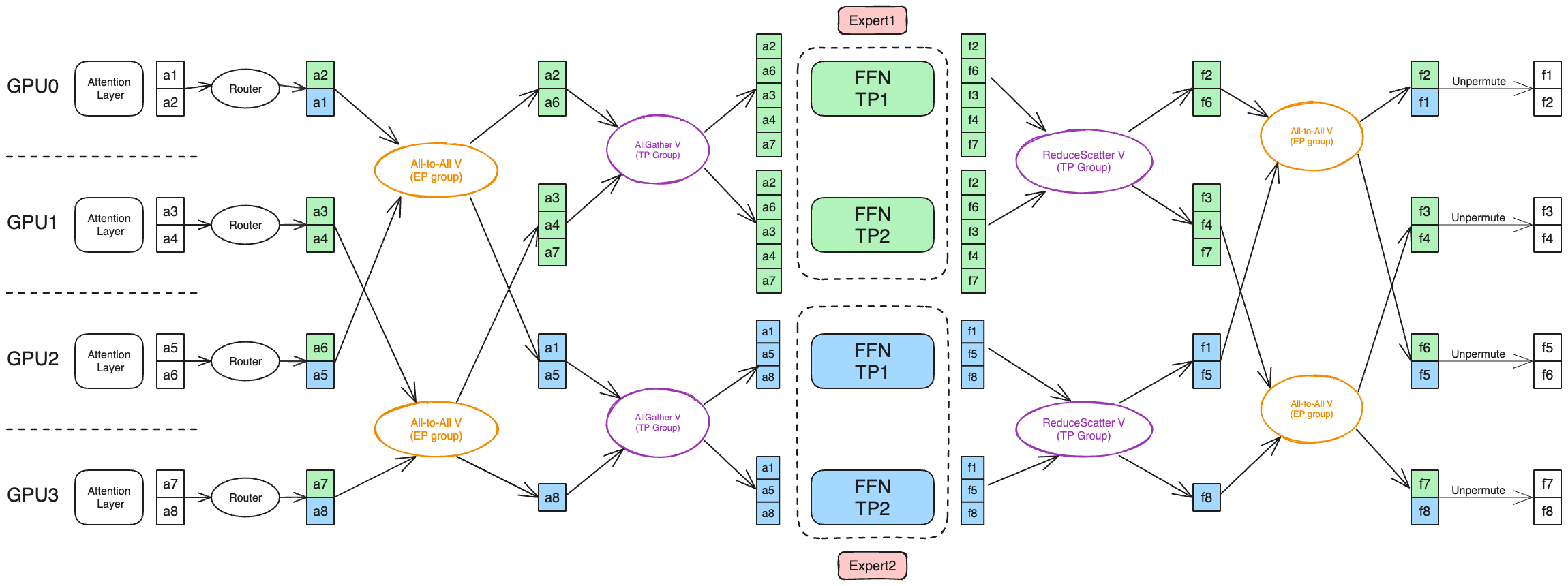}
    \caption{Workflow of token dispatcher with Tensor Parallelism and Expert
    Parallelism.}
    \label{fig:token-dispatcher}
  \end{figure}

  % Parallelism mappings
  In Figure~\ref{fig:token-dispatcher}, we illustrate the workflow of an MoE layer distributed
  across four GPUs, where the degrees of TP and ETP are both 2. GPU pairs (0, 1) and
  (2, 3) form the ETP group. GPU pairs (0, 2) and (1, 3) form the EP group.

  % Forward workflow
  The forward computation workflow proceeds as follows. First, the router determines the mapping of each token to its designated expert based on the local input and reorganizes the tokens assigned to the same expert into contiguous memory regions through a permutation operation. Next, an All-to-All-V communication is executed across the EP groups to exchange tokens, ensuring that each token is delivered to its corresponding expert. Following this, an AllGather-V communication is performed within the ETP groups to guarantee that all members within an ETP group share identical activations. Once the AllGather-V communication is complete, each GPU computes its assigned partition of the expert feed-forward networks. A subsequent ReduceScatter-V communication within the ETP groups aggregates and distributes the output hidden states, effectively reversing the AllGather operation. Another All-to-All-V communication is then employed to return the tokens to their original GPUs. Finally, an un-permutation operation restores the tokens to their initial order, preparing them for further processing in the attention layer. The backward workflow mirrors the forward process, with the AllGather/ReduceScatter (AG/RS) operations in the TP groups replaced by ReduceScatter/AllGather (RS/AG).

  We now elaborate on the design of the router to support both token-dropping and token-dropless training paradigms. The router assigns tokens to experts by selecting the top-$k$ tokens based on their softmax probabilities. In token-dropless training, ensuring numerical correctness is straightforward, as token assignments remain consistent across different parallelism configurations. For token-dropping training, two potential strategies can be employed: \textbf{full-sequence-based dropping} and \textbf{sub-sequence-based dropping}.
  
\begin{itemize}
    \item \textbf{Full-sequence dropping} ensures consistency by gathering logits from all ranks that collectively represent the entire sequence. However, this approach incurs significant communication overhead, particularly when sequences are distributed across multiple nodes.
    
    \item \textbf{Sub-sequence dropping}, on the other hand, makes dropping decisions based solely on the logits from the current sub-sequence. This strategy eliminates the need for gathering logits across ranks, thereby reducing communication overhead and alleviating load imbalance issues during token communication.
\end{itemize}

Empirically, we observe that sub-sequence dropping does not adversely affect model convergence compared to full-sequence dropping. Consequently, we adopt the sub-sequence dropping approach as the default strategy in this work.

  \section{Experiments}

  \subsection{Experimental Setup}

  All experiments in this work were conducted on the Eos~\cite{Eos} cluster. 
  The Eos cluster consists of NVIDIA DGX H100 nodes, each equipped with eight NVIDIA H100 GPUs~\cite{H100} 
  and two 56-core Intel Sapphire Rapids CPUs. Each GPU achieves a peak half-precision throughput of 989.5
  TFLOP/s, and all GPUs are interconnected via NVLink 4th Generation~\cite{NVLINK}
  and InfiniBand~\cite{InfiniBand}. The peak uni-directional communication bandwidths are 450 GB/s
  for intra-node (NVLink) and 400Gbps for inter-node (InfiniBand) connections.
  We utilize PyTorch 2.5.0 and CUDA 12.6 for our experiments. 
  All performance measurements reported in TFLOPS and MFU are conducted using BF16 precision.
  Up to 1024 GPUs are utilized in the scaling experiments.

  We select two types of MoE models for our experiments, coarse-grained and fine-grained
  MoE, each type containing models of two different sizes. Compared to coarse-grained
  MoE, fine-grained MoE has a larger number of experts and more activated experts per
  token, but each expert has a reduced hidden size. For the coarse-grained MoE,
  we select the Mixtral 8x22B~\cite{mixtral-8x22b} model and design a larger MoE
  named Llama3-8x70B by upcycling Llama3-70B~\cite{llama3} to 8 experts~\cite{upcycling}.
  For the fine-grained MoE, we choose Qwen2-57B-A14B~\cite{qwen2}, which has 64 experts and 8 active experts per token, totaling 57 billion parameters with 14 billion
  active parameters. To obtain a larger fine-grained MoE model, we
  reparameterized the Mixtral 8x22B model to 64 experts and 8 active experts per
  token called Mixtral-8x22B-G8T8, with each expert possessing a hidden size that is one-eighth of the original model, by applying fine-grained upcycling~\cite{FineGrained}.

  \subsection{Performance Comparison}
  To evaluate the performance of our proposed MoE Parallel Folding technique
  compared to existing parallelism strategies, we conducted comparative experiments
  using the four models previously described. The primary metric for assessment
  was the Model TFLOPS Utilization (MFU) during training, which measures the efficiency
  of computational resource utilization by comparing theoretical peak performance
  with the actual achieved performance in BF16 precision. 
  To alleviate the performance jitter caused by load imbalance issues in dropless training, we use token drop training with a capacity factor equal to 1 for benchmarking.

  For baseline comparisons, we chose four representative baseline parallelism
  strategies:
  \begin{enumerate}
    \item FSDP~\cite{fsdp}: A data parallelism method that shards model
      parameters, gradients, and optimizer states across workers.

    \item FSDP + EP~\cite{megablocks}: An extension of FSDP that incorporates EP.

    \item TP+EP+DP \cite{TED}: An framework combining TP and EP to fit larger
      MoE models across multiple GPUs.

    \item MCore with 5D-parallelism\cite{Megatron-3d}: The state-of-the-art training
      framework for large scale LLM models, supporting TP,EP,CP,DP and PP.
  \end{enumerate}
  All baseline methods were implemented using the NVIDIA Megatron-Core framework\footnote{\url{https://github.com/NVIDIA/Megatron-LM}}.
  For each method, we report the MFU achieved with the optimal parallelism configuration found by tuning its supported parallelism dimensions.

   \begin{table}[h!]
    \centering
    \caption{Performance comparison of different parallelism strategies by MFU. The global batch size for experiment is 256.}
    \begin{tabular}{ccccc}
        \toprule
        & \multicolumn{2}{c}{Coarse-grained} & \multicolumn{2}{c}{Fine-grained} \\ 
        \midrule
        & Mixtral-8x22B & Llama3-8x70B & Qwen2-57B-A14B & Mixtral-8x22b-G8T8 \\ 
        GPUs & 128 & 256 & 64 & 128 \\ 
        \midrule
        FSDP & 4.3\% & OOM & 9.9\% & 2.2\% \\ 
        FSDP + EP & 23.4\% & 19.6\% & 25.4\% & 9.0\% \\ 
        TP+EP+DP & 36.6\% & OOM & 23.1\% & 8.7\% \\ 
        MCore & 46.3\% & 38.8\% & 35.3\% & 17.1\% \\ 
        MCore w/ Folding & \textbf{49.3\%} & \textbf{41.6\%} & \textbf{39.0\%} & \textbf{28.8\%} \\ 
        \bottomrule
    \end{tabular}
    \label{tab:performance_comparison}
\end{table}

  Table~\ref{tab:performance_comparison} presents the comparison results of
  different parallelism strategies on the selected MoE models. The observed MFU values
  highlight several key insights into the performance implications of each strategy:
  (1) \textbf{FSDP} exhibits poor performance(<10\% MFU) due to their sparse computations and large parameter counts. In FSDP, the
  communication of parameters and gradients cannot be effectively overlapped
  with computation. Additionally, FSDP fails to train larger models like Llama3-8x70B
  due to out-of-memory (OOM) issues. (2) \textbf{FSDP + EP} improves performance,
  by parallelizing expert across GPUs, thereby reducing communication of expert parameters and gradients. However, this strategy still suffers from
  communication overhead that cannot be fully overlapped with computation, limiting further
  performance gains. (3)\textbf{TP + EP + DP}~\cite{TED} further uses TP to
  split the model weights to multiple GPUs and use ZeRO-1 instead of ZeRO-3 to
  reduce communication overhead of parameters, resulting in better performance.
  But a large TP also introduces significant activation communication overhead. And the largest model Llama3-8x70B could not be trained using only TP+EP due to memory constraints. (4)\textbf{MCore}
  framework leverages pipeline parallelism (PP) in addition to TP, EP, and DP, achieves
  a better balance between communication and computation. This results in higher MFU values, reaching
  46.3\% on Mixtral-8x22B and 35.3\% on Qwen-2-57B. By effectively
  partitioning the model across pipeline stages, MCore reduces the memory
  footprint per GPU and overlaps communication with computation more efficiently.
  However, the coupling of parallelism strategies between the Attention and MoE layers renders the mappings sub-optimal for MoE models. (5)\textbf{MCore with MoE
  Parallel Folding:} further enhances training efficiency, achieving the highest
  MFU values across all models: 49.3\% for Mixtral-8x22B, 41.6\% on Llama3-8x70B,
  39.0\% on Qwen-2-57B, and 28.8\% on Mixtral-8x22B-G8T8. The flexible
  parallelism provided by MoE Parallel Folding allows for a more optimal
  parallelism strategy tailored to the characteristics of MoE models. By folding
  MoE parallel groups with Attention and effectively utilizing available hardware
  resources, it minimizes communication overhead and maximizes computational
  efficiency. This leads to significant performance improvements over existing strategies.

  \begin{figure}[t]
    \centering
    \includegraphics[width=\textwidth]{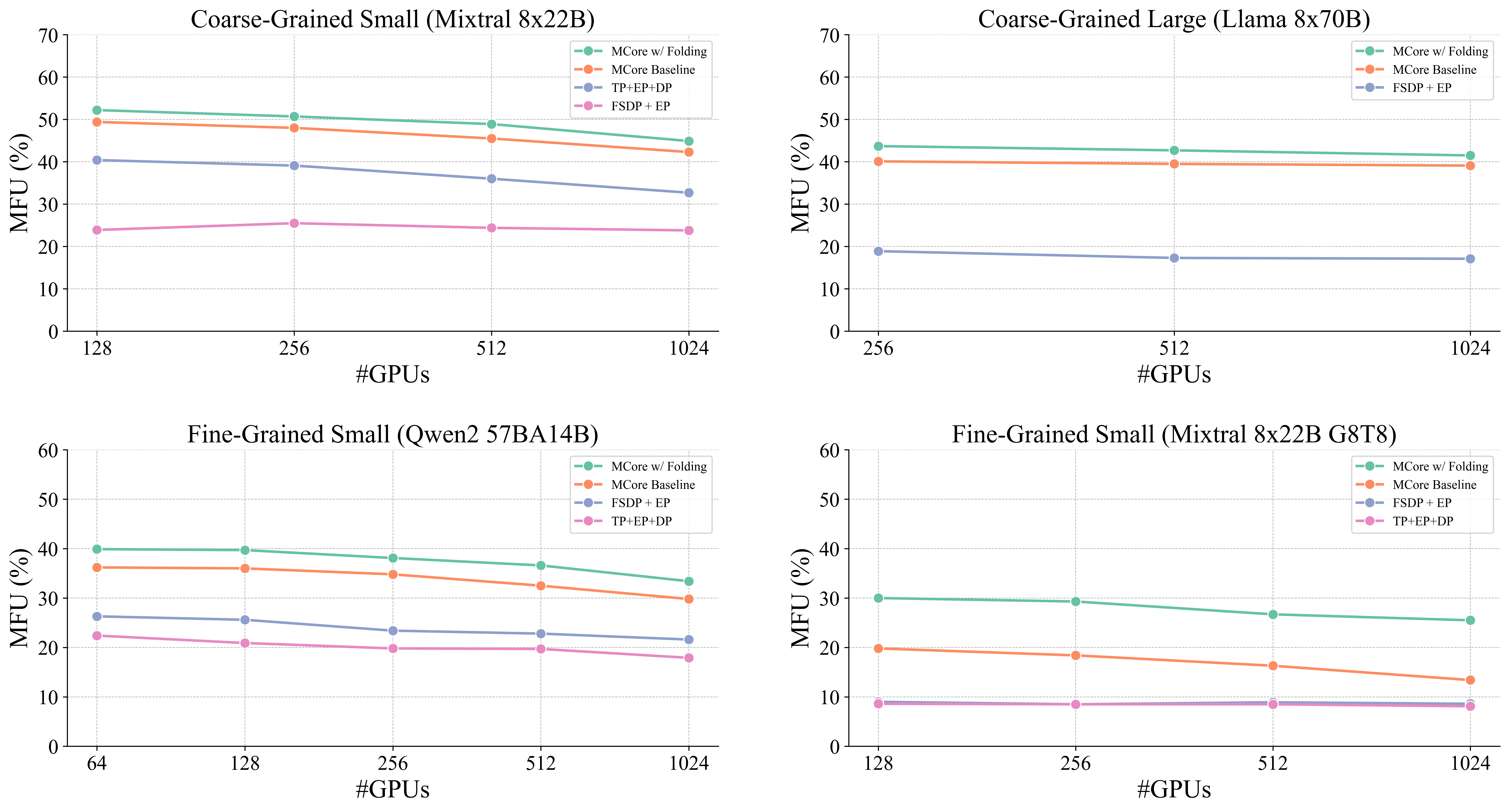}
    \caption{Strong scaling experiments for various parallelism strategies by
    increasing number of GPUs up to 1024.}
    \label{fig:strong-scaling}
  \end{figure}

  % Context scaling

  \begin{figure}[b]
    \centering
    \includegraphics[width=\textwidth]{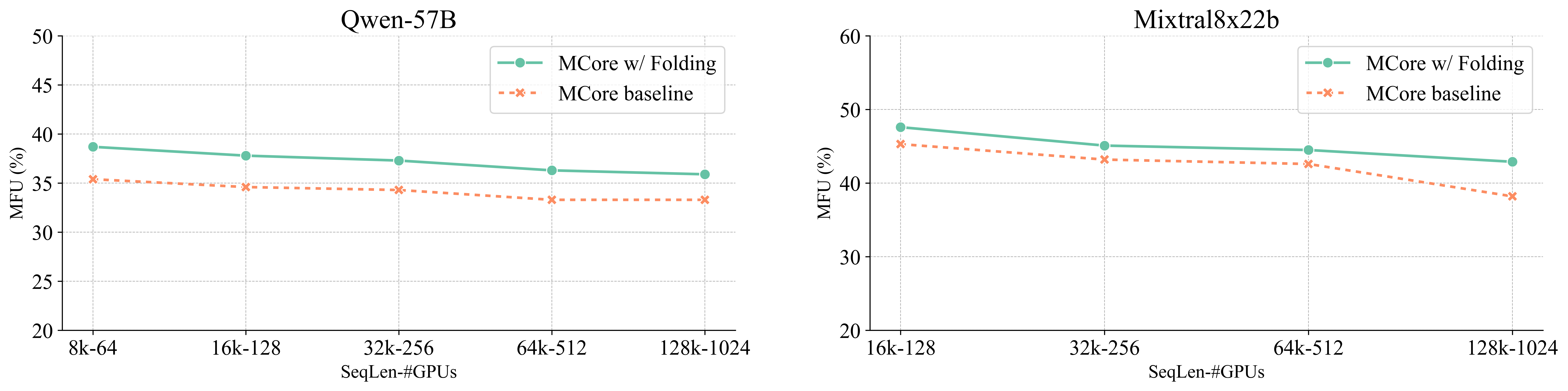}
    \caption{Context-scaling experiments by increasing context length and number of GPUs up to 128K and 1024.}
    \label{fig:context-scaling}
  \end{figure}

  The experiments also reveal that fine-grained MoE models achieve lower training efficiency compared to coarse-grained MoE models across all parallelism strategies. This performance gap stems from two key factors: (1) Fine-grained MoE models
  generate higher communication volume due to their architecture - they employ
  more experts and activate more experts per token, increasing communication overhead during the token dispatching process. 
  Additionally, the smaller hidden sizes decrease GEMM efficiency.
  (2) Fine-grained MoE models typically incorporate a larger number of local and active experts, leading to significant memory overhead for storing activations. The memory requirements for managing numerous experts force the use of larger model parallelism sizes, which introduces additional communication costs and further reduces training efficiency.

  \subsection{Scaling Experiments}
  \paragraph{Strong Scaling}
  To evaluate the scalability of our methods, we conduct strong scaling experiments by increasing the number of GPUs up to 1,024.
  The global batch size is set to 1024 in the scaling experiments.
  As shown in Figure~\ref{fig:strong-scaling}, our framework maintains consistently higher MFU compared to baseline approaches as the GPU count increases across all model types.
  The results show the scalability of MoE parallel folding up to 16x nodes with little MFU drops, especially for large-scale models like Llama3-8x70B, where the MFU only drops from 43.7\% to 41.5\%.

  \paragraph{Scaling with Context Length}
  To evaluate the capability of our framework to train large scale MoE models with very long context lengths, we conducted context scaling experiments by increasing the sequence length while keeping the total number of tokens per global batch constant.
  As shown in Figure~\ref{fig:context-scaling}, our framework can train MoE models with high efficiency up to a context length of 128K tokens, and the MFU only drops from 38.7\% to 35.9\% for Qwen-57B14A and 47.6\% to 42.9\% for Mixtral-8x22B.
  With MoE parallel folding, MCore can achieve higher performance by folding the parallelism groups of attention and MoE layers to better utilize the intra-node communication bandwidth.

  % CP2EP8 w/o folding -> CP2EP8 w folding
  % \begin{figure}[h]
  %   \vspace{-10pt}
  %   \centering
  %   \includegraphics[width=\textwidth]{images/ablation-8x22B.png}
  %   \caption{MoE layer breakdown for Mixtral 8x22B model with different parallelism mappings. Marker * means the new parallelism mappings supported by MoE Parallel Folding.}
  %   \label{fig:ablation-8x22b}
  % \end{figure}

  % \begin{figure}[h]
  %   % \vspace{-20pt}
  %   \centering
  %   \includegraphics[width=\textwidth]{images/ablation-fine-grained.png}
  %   \caption{MoE layer breakdown for Mixtral 8x22B G8T8 model with different parallelism mappings. Marker * means the new parallelism mappings supported by MoE Parallel Folding.}
  %   \label{fig:ablation-8x22bg8t8}
  % \end{figure}

\subsection{Ablation Study}
\begin{figure}
    \begin{subfigure}[b]{0.49\textwidth}
      \centering
      \includegraphics[width=\textwidth]{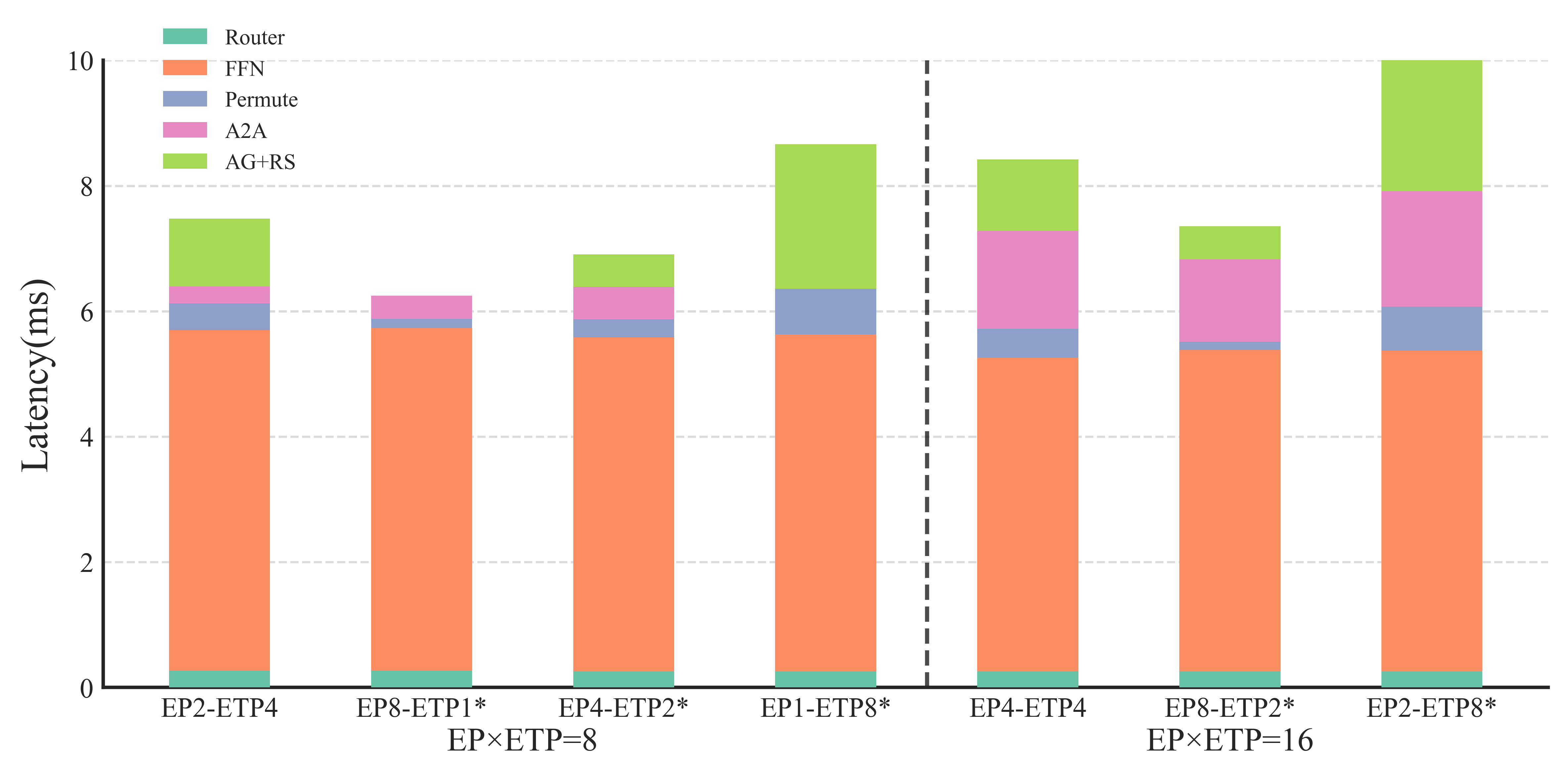}
      \caption{Mixtral 8x22B model}
      \label{fig:ablation-8x22b}
    \end{subfigure}
    \hfill
    \begin{subfigure}[b]{0.49\textwidth}
      \centering
      \includegraphics[width=\textwidth]{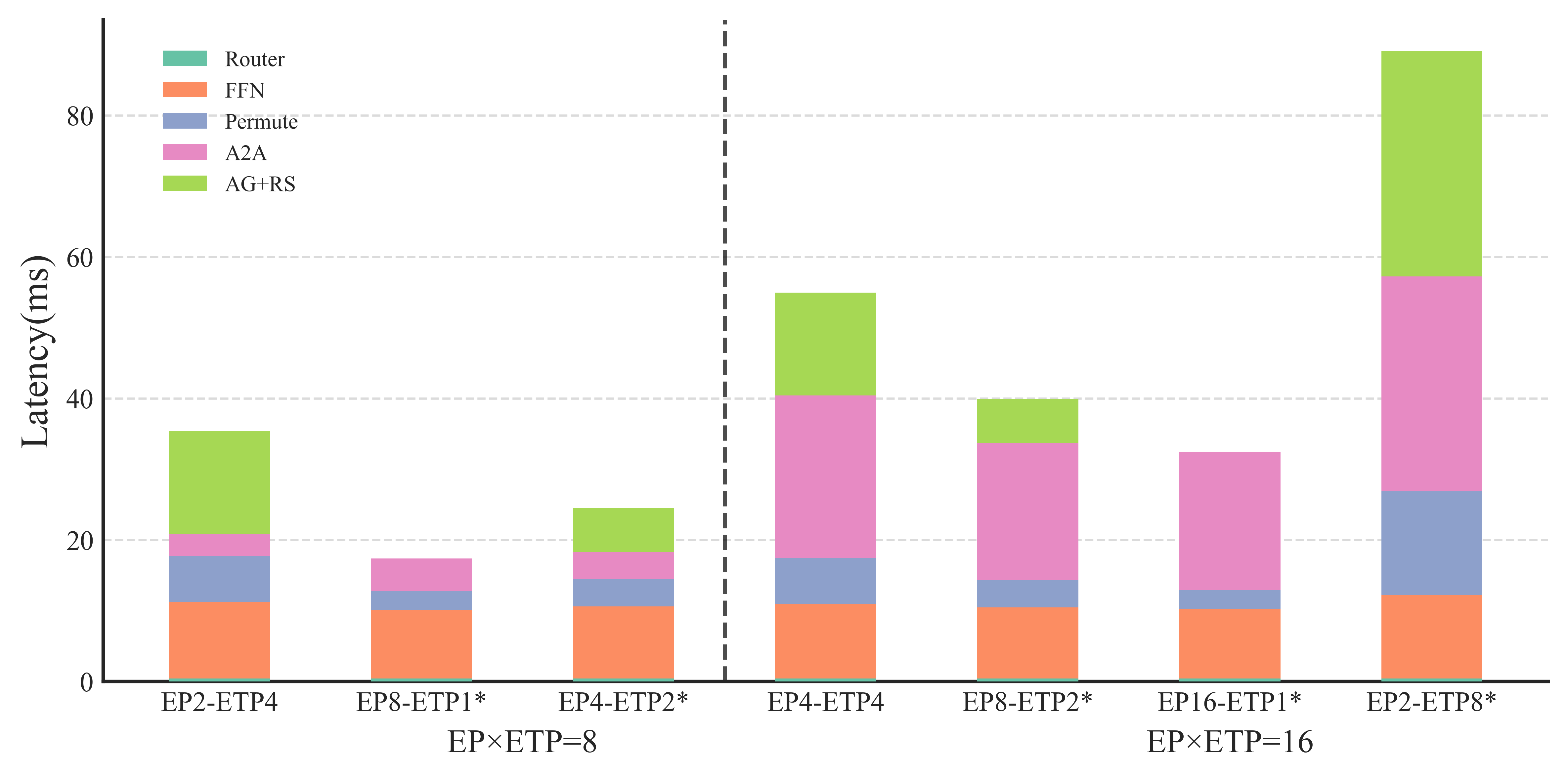}
      \caption{Mixtral 8x22B G8T8 model}
      \label{fig:ablation-8x22bg8t8}
    \end{subfigure}
    \caption{MoE layer breakdown with different parallelism mappings. Marker * means the new parallelism mappings supported by MoE Parallel Folding.}
    \label{fig:ablation-ETP}
\end{figure}

  To systematically evaluate the performance characteristics of MoE layers and quantitatively assess the advantages of MoE parallel folding, we conducted comprehensive ablation studies. Our methodology involves varying the parallelism mappings of the MoE layer while maintaining fixed parallelism configurations for the Attention layer. 
  Specifically, we examine the Attention layer's parallelism mappings across TP and CP, while the MoE layer's parallelism mappings are analyzed with respect to EP and ETP.

  In the first experimental setup, we configure the Attention layer with TP=4 and CP=1 (no context parallelism). We evaluate parallelism mappings for the MoE layer with EPxETP=8 and EPxETP=16, which enables us to examine both intra-node and inter-node communication patterns. Notably, the memory utilization remains consistent across different configurations when the product ETPxEP is held constant.

  Figure~\ref{fig:ablation-ETP} presents detailed latency breakdowns for the MoE layer in both the standard Mixtral 8x22B model and its fine-grained variant Mixtral 8x22B G8T8. Configurations enabled by MoE parallel folding are denoted with an asterisk (*). Our analysis reveals several key findings:
  (1) MoE Parallel Folding significantly expands the available parallelism configuration space, enabling the discovery of optimal parallelism mappings. The configurations utilizing MoE parallel folding consistently achieve superior performance.
  (2) ETP in the MoE layer introduces substantially higher communication overhead compared to EP, with this effect being particularly pronounced in fine-grained MoE models.
  (3) Fine-grained MoE models exhibit notably lower computation-to-communication ratios. When ETPxEP exceeds 8, necessitating inter-node communication, communication overhead dominates, accounting for over 70\% of the total latency.
  (4) Maintaining minimal model parallelism while favoring EP over ETP emerges as an effective strategy for optimizing MoE layer performance.

\begin{figure}
    \begin{subfigure}[b]{0.49\textwidth}
      \centering
      \includegraphics[width=\textwidth]{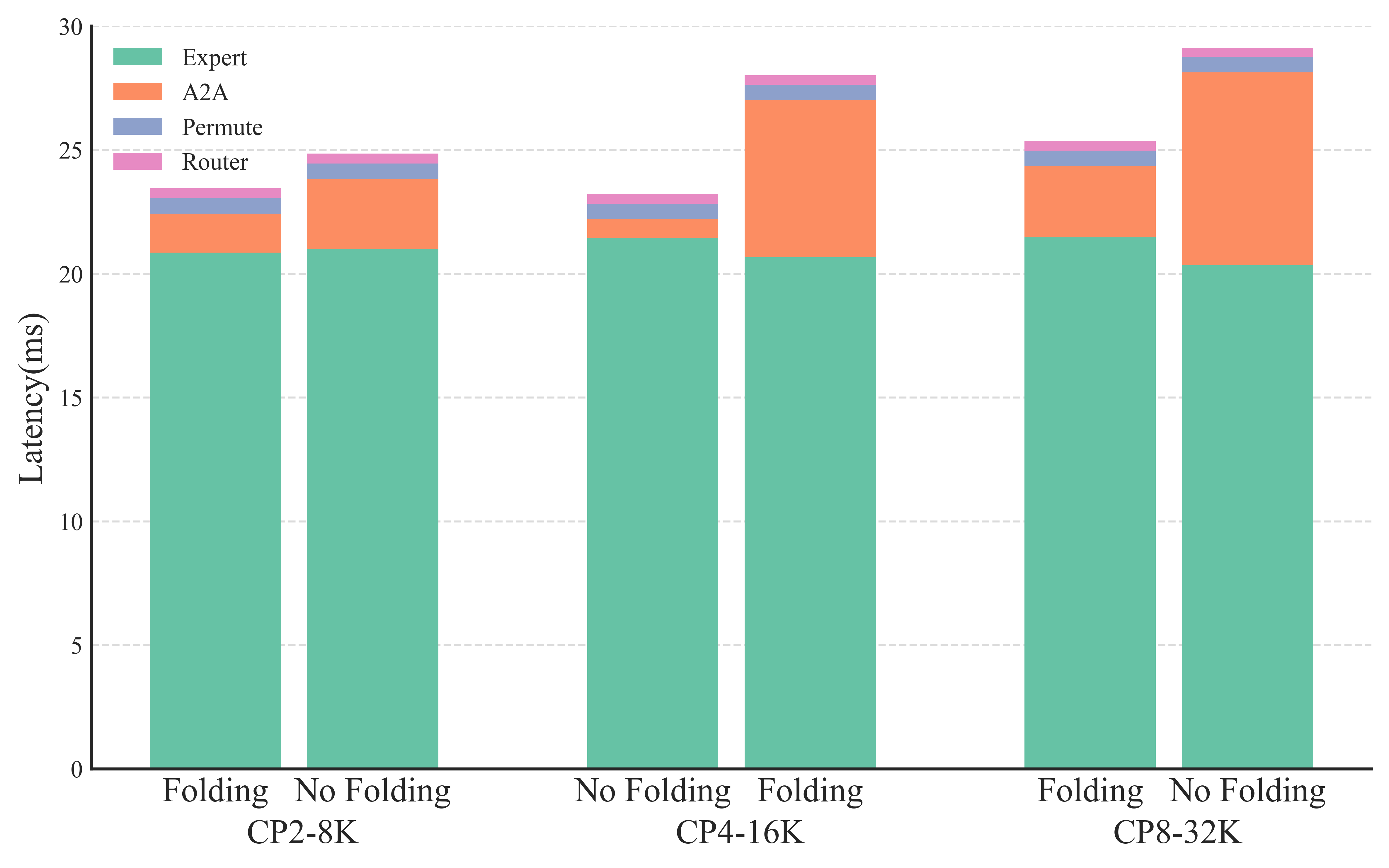}
      \caption{Mixtral 8x22B model}
      \label{fig:ablation-8x22B-context}
    \end{subfigure}
    \hfill
    \begin{subfigure}[b]{0.49\textwidth}
      \centering
      \includegraphics[width=\textwidth]{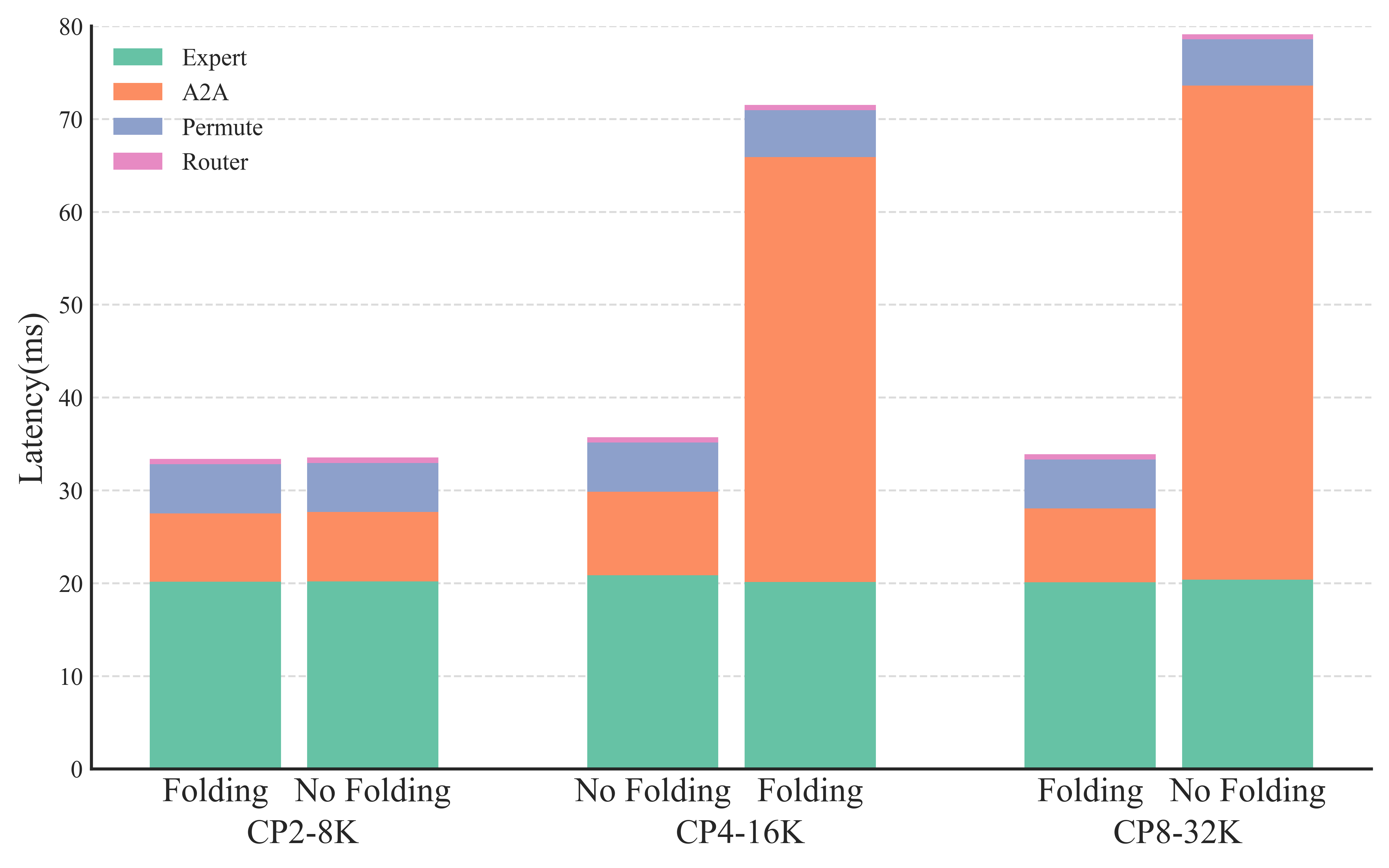}
      \caption{Mixtral 8x22B G8T8 model}
      \label{fig:ablation-8x22bg8t8-context}
    \end{subfigure}
    \caption{MoE layer breakdown with different parallelism mappings. Marker * means the new parallelism mappings supported by MoE Parallel Folding.}
    \label{fig:ablation-ECP}
\end{figure}

  In the second experimental setup, we configure the Attention layer with various CP sizes and sequence lengths, and compare the performance of the MoE layer with and without parallel folding. Figure~\ref{fig:ablation-ECP} shows the breakdown results. As we can see, when the size of the CPxEP group exceeds 8 and spans beyond the NVLINK domain,
  the latency without MoE Parallel Folding increases significantly. 
  Without MoE Parallel Folding, the EP group spans across multiple context parallelism groups, causing All-to-All communications within the EP group to traverse the lower-bandwidth inter-node network fabric. The MoE Parallel Folding technique allows the CP and EP groups to be folded together, maximizing the use of high-bandwidth NVLink connections whenever possible.

\subsection{FP8 Training Performance}
To further evaluate the capabilities of our framework, we investigated the performance benefits of utilizing FP8 precision, particularly relevant for newer hardware architectures like NVIDIA Hopper and NVIDIA Blackwell. We conducted experiments employing FP8 delayed scaling~\cite{fp8_primer} with the Mixtral 8x22B model on 128 H100 GPUs. The results demonstrate substantial throughput improvements compared to BF16 training.

Specifically, we observed the following performance in model TFLOPS:
\begin{table}[h]
  \centering
  \caption{Mixtral 8x22B Performance Comparison}
  \label{tab:fp8_performance}
  \begin{tabular}{ccccc}
    \toprule
    Configuration & Precision & TFLOPS & Speedup vs BF16 & Speedup w/ Folding \\ \midrule
    MCore & BF16 & 458.3 & - & - \\
    MCore w/ Folding & BF16 & 487.7 & - & 1.06x \\
    MCore & FP8 & 575.1 & 1.26x & - \\
    MCore w/ Folding & FP8 & 631.7 & 1.30x & 1.10x \\
    \bottomrule
  \end{tabular}
\end{table}

These results, summarized in Table~\ref{tab:fp8_performance}, indicate that FP8 training provides a significant performance uplift over BF16 (approximately 1.26x speedup without folding and 1.30x with folding). Furthermore, MoE Parallel Folding continues to enhance performance within the FP8 regime, yielding the highest throughput of 631.7 TFLOPS.

\section{Conclusion}
In this paper, we introduce a novel framework for efficient large-scale MoE model training that addresses key challenges in distributed training through two main innovations. 
First, we propose MoE Parallel Folding, a technique that decouples the parallelization strategies of attention and MoE layers, 
enabling more flexible and efficient parallel configurations. This approach allows for optimal resource utilization by adapting to the distinct computational characteristics of each layer. Second, we develop an efficient token-level dispatcher that supports both token-dropping and token-dropless training across five dimensions of parallelism, providing a robust foundation for complex hybrid parallelism schemes.
Our experimental results demonstrate significant performance improvements across different MoE architectures, achieving up to 49.3\% MFU for Mixtral 8x22B and 39.0\% MFU for Qwen2-57B-A14B on H100 GPUs. The framework shows strong scaling efficiency up to 1024 GPUs and maintains high performance with sequence lengths up to 128K tokens. These results validate the effectiveness of our approach in addressing the scalability challenges of large-scale MoE model training.

\newpage
\bibliographystyle{plain}
\bibliography{bibliography}

  \newpage

  \section{Appendix}
  \subsection{Accuracy Validation}
\begin{figure}[h]
    \vspace{-10pt}
    \centering
    \includegraphics[width=\textwidth]{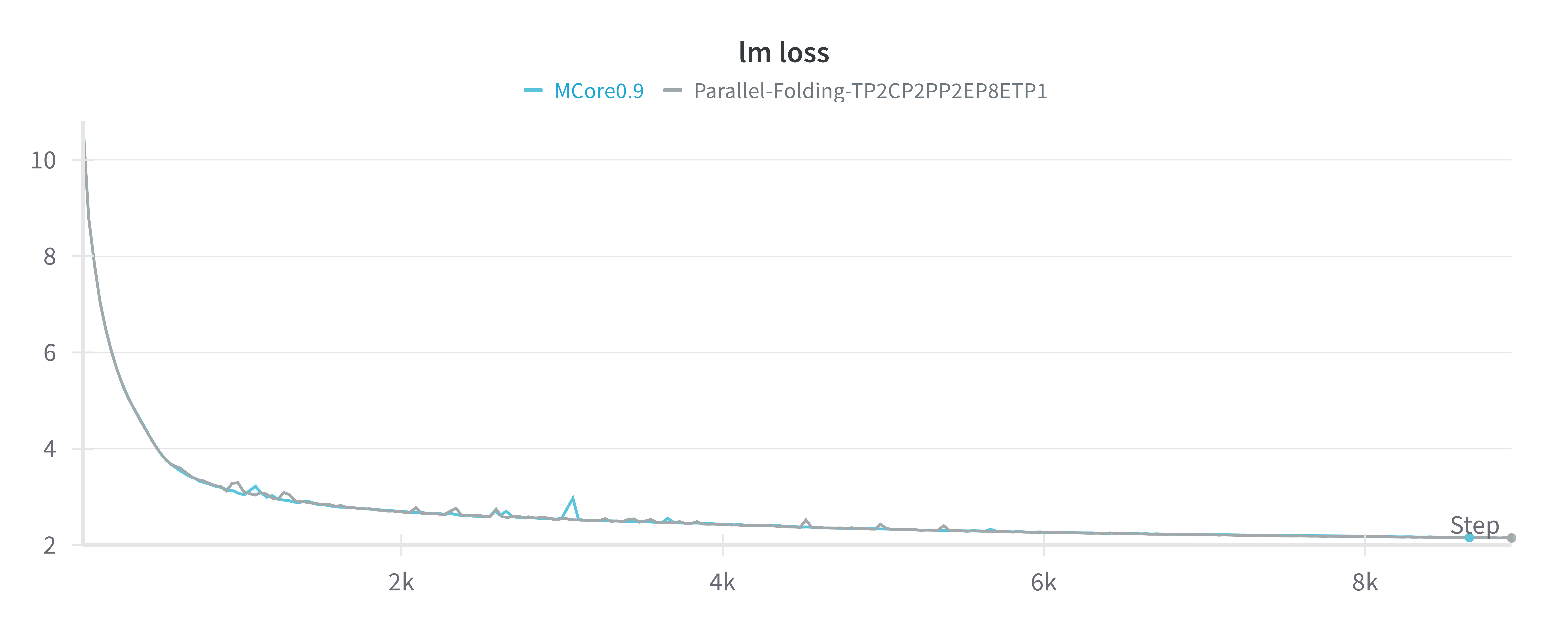}
    \caption{Training loss of MCore with MoE Parallel Folding compared to MCore v0.9.}
    \label{fig:train-loss}
  \end{figure}

\begin{figure}[h]
    % \vspace{-20pt}
    \centering
    \includegraphics[width=\textwidth]{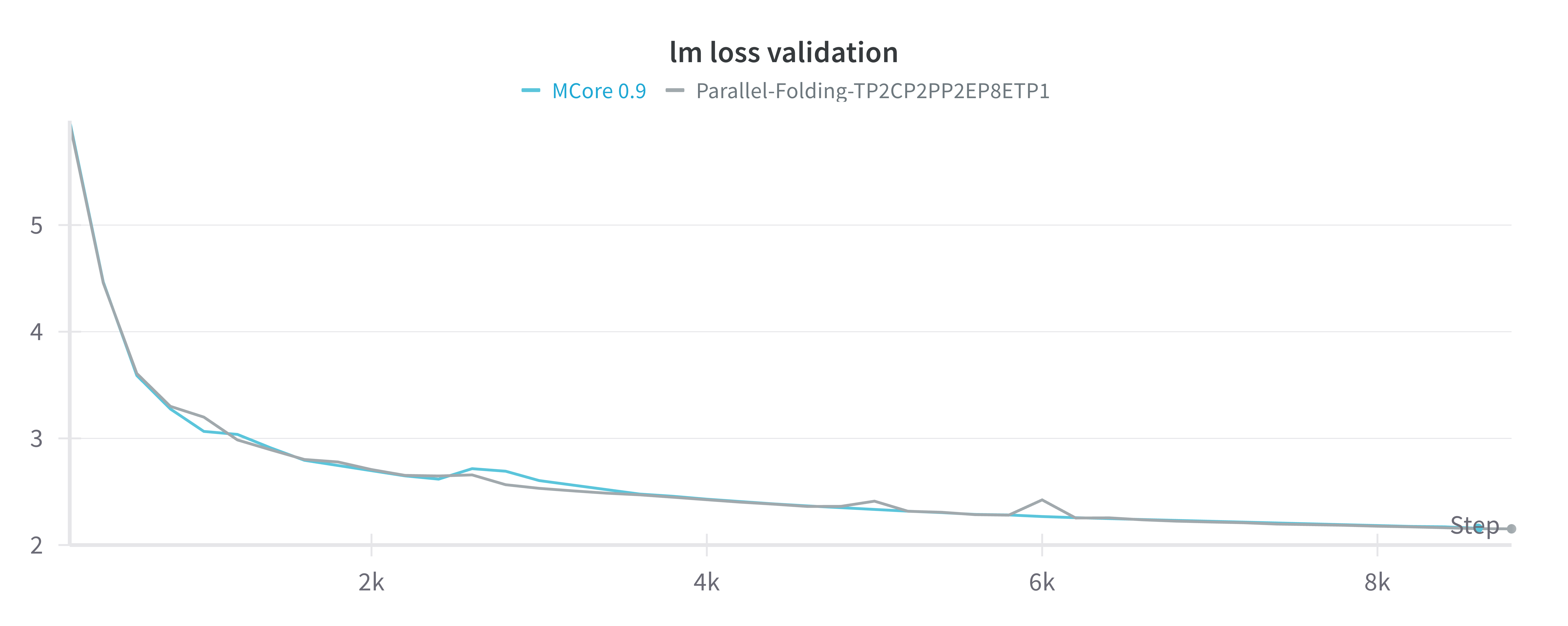}
    \caption{Validation loss of MCore with MoE Parallel Folding compared to MCore v0.9.}
    \label{fig:valid-loss}
  \end{figure}

  To validate the accuracy of our implementation, we train a Mixtral 8x7B model with MoE Parallel Folding compared to MCore v0.9 in a token-dropless manner up to 40B tokens.
  We set TP=2, CP=2, PP=2, EP=8, ETP=1, this allows us to verify the correctness of MoE Parallel Folding where EP in MoE layer are folded with all of TP,CP,DP in Attention.
  As shown in Figures~\ref{fig:train-loss} and~\ref{fig:valid-loss}, MCore with MoE Parallel Folding is able to successfully train the model to convergence, and the training and validation loss curves align well with MCore v0.9.

  \newpage
  \subsection{Workflow for Transformer Layer with MoE parallel folding}

\begin{figure}[h]
    % \vspace{-20pt}
    \centering
    \includegraphics[width=\textwidth]{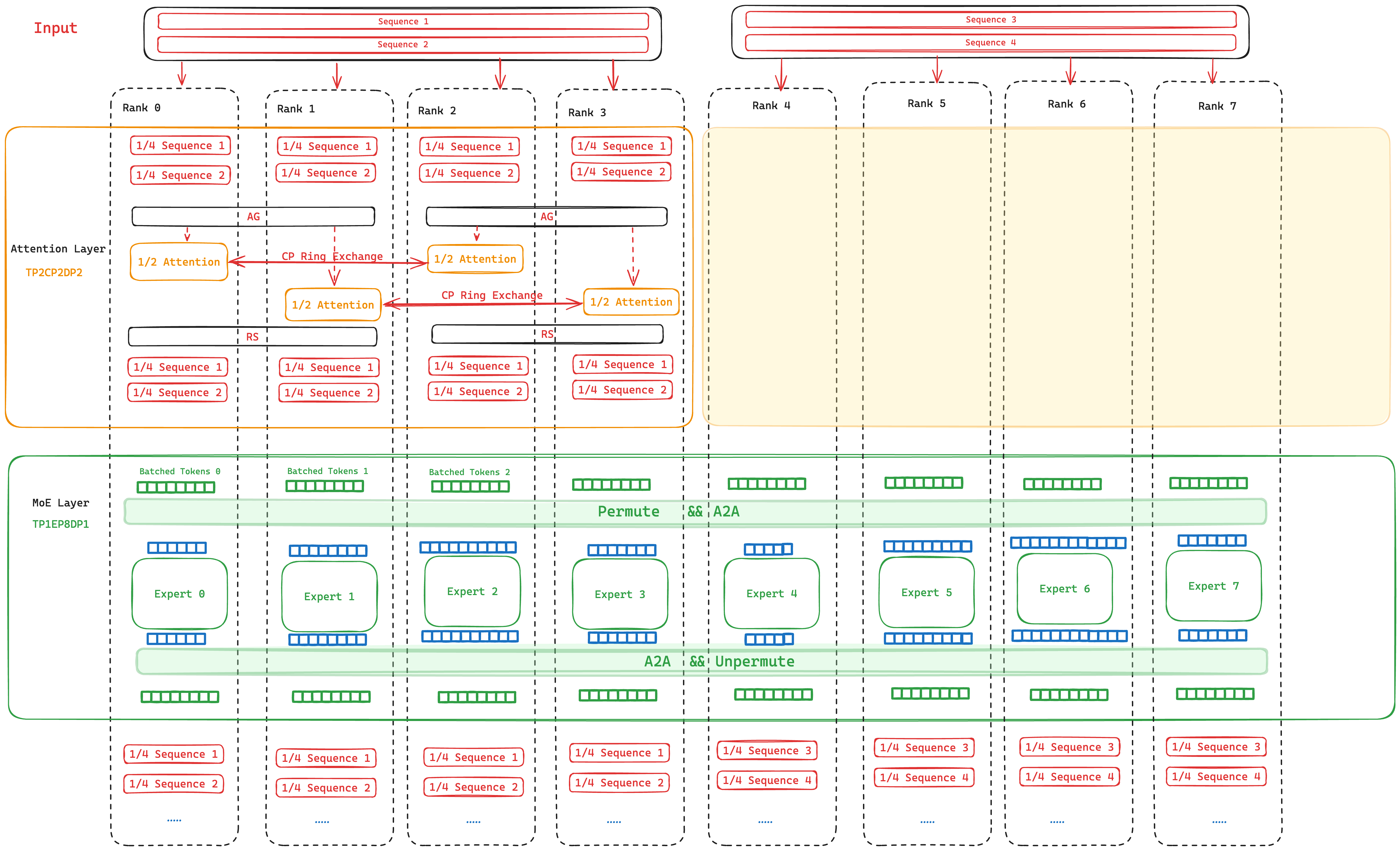}
    \caption{Workflow of the Transformer layer with MoE Parallel Folding.}
    \label{fig:workflow}
  \end{figure}

  Figure~\ref{fig:workflow} illustrates the overall workflow of a Transformer Layer in an MoE model with Parallel Folding.
  In the Attention component, the parallelism mapping is TP2CP2DP2, where each sequence is split across 4 GPUs. For the MoE layer, the parallelism mapping is TP1EP8DP1, with each GPU handling a different expert FFN.
  The transformation between the Attention and MoE layers requires only a reshape operation that flattens the sequence/subsequence into a batch of tokens, introducing no explicit communication overhead.

  \newpage
  \subsection{Parallel Groups Initialization}
  In the code of Listing~\ref{lst:parallel-groups}, we give an example to show how the parallelism groups with MoE Parallel Folding for each device are generated. The code demonstrates the initialization of parallel groups for both attention and MoE components, handling different parallelism dimensions including TP, EP, PP and DP.

  The function \texttt{generate\_mappings} takes the total number of devices (\texttt{world\_size}) and parallelism dimensions as input parameters. It first calculates the effective data parallelism degrees for attention and MoE components separately. Then, it creates two sets of parallel groups: one for attention layers with TP, CP, PP, and DP dimensions, and another for MoE layers with TP, EP, PP, and DP dimensions.

  \begin{lstlisting}[
    language=Python,
    basicstyle=\footnotesize \ttfamily,
    caption={Python implementation of parallel group generation for MoE Parallel Folding},
    label={lst:parallel-groups}
]
from einops import rearrange
import torch

def generate_mappings(world_size, tp, cp, ep, etp, pp):
    ranks = torch.arange(world_size)
    attn_dp = world_size // tp // cp // pp
    moe_dp = world_size // etp // ep // pp

    # Parallel groups for attention
    attn_ranks = ranks.reshape(attn_dp, pp, cp, tp)
    attention_groups = {
        "TP": rearrange(attn_ranks, "attn_dp pp cp tp -> (attn_dp pp cp) tp", 
                       tp=tp, cp=cp, pp=pp, attn_dp=attn_dp).tolist(),
        "CP": rearrange(attn_ranks, "attn_dp pp cp tp -> (attn_dp pp tp) cp",
                       tp=tp, cp=cp, pp=pp, attn_dp=attn_dp).tolist(),
        "PP": rearrange(attn_ranks, "attn_dp pp cp tp -> (attn_dp cp tp) pp",
                       tp=tp, cp=cp, pp=pp, attn_dp=attn_dp).tolist(),
        "DP": rearrange(attn_ranks, "attn_dp pp cp tp -> (pp cp tp) attn_dp",
                       tp=tp, cp=cp, pp=pp, attn_dp=attn_dp).tolist()
    }

    # Parallel groups for MoE
    moe_ranks = ranks.reshape(moe_dp, pp, ep, tp)
    moe_groups = {
        "TP": rearrange(moe_ranks, "moe_dp pp ep tp -> (moe_dp pp ep) tp",
                       tp=etp, ep=ep, pp=pp, moe_dp=moe_dp).tolist(),
        "EP": rearrange(moe_ranks, "moe_dp pp ep tp -> (moe_dp pp tp) ep",
                       tp=etp, ep=ep, pp=pp, moe_dp=moe_dp).tolist(),
        "PP": rearrange(moe_ranks, "moe_dp pp ep tp -> (moe_dp ep tp) pp",
                       tp=etp, ep=ep, pp=pp, moe_dp=moe_dp).tolist(),
        "DP": rearrange(moe_ranks, "moe_dp pp ep tp -> (pp ep tp) moe_dp",
                       tp=etp, ep=ep, pp=pp, moe_dp=moe_dp).tolist()
    }

    return attention_groups, moe_groups

attn_groups, moe_groups = generate_mappings(64, 2, 2, 2, 2, 2)
\end{lstlisting}

\subsection{Details of Parallelism Mappings in Experiments}
We conducted numerous experiments to find the optimal training parallel configurations. The optimal settings found and their corresponding performance metrics are presented in Table \ref{tab:detailed_perf_comparison}. In these experiments, the global batch size was fixed at 256, and the sequence length was fixed at 4096.

To investigate the scalability of various methods, we fixed the parallel configuration and increased the number of GPUs. The detailed benchmark numbers are presented in Table \ref{tab:detailed_GPUs_scaling}. All parallel configurations are the same as those identified in the performance experiments. 

In the context scaling experiment, influenced by the long sequence length, the optimal parallel configuration might differ. The parallel configurations found and the detailed performance results are presented in Table \ref{tab:detailed_context_scaling}.

\begin{table}[h!]
    \centering
    \caption{Detailed parallel mapping of models with optimal configurations.}
    \label{tab:detailed_perf_comparison}
    \begin{tabular}{l l c c c c c c c}
        \toprule
        Model & Methods & GPUs & CP & TP & EP & PP & ETP & MFU \\ 
        \midrule
        \multirow{5}{*}{Mixtral-8x22B} 
        & FSDP & 128 & 1 & 8 &  &  &  & 4.3\% \\ 
        & FSDP + EP & 128 & 1 & 2 & 8 &  &  & 23.4\%  \\ 
        & TP + EP + DP & 128 & 1 & 4 & 8 &  &  & 36.6\%  \\ 
        & MCore & 128 & 1 & 2 & 4 & 8 &  & 46.3\% \\ 
        & MCore w/ Folding & 128 & 1 & 2 & 8 & 8 & 1 & 49.3\% \\ 
        \midrule
        \multirow{5}{*}{Qwen2-57B-A14B} 
        & FSDP & 64 & 1 & 2 & 1 &  &  & 9.9\% \\ 
        & FSDP + EP & 64 & 1 & 1 & 8 &  &  & 25.4\% \\ 
        & TP + EP + DP & 64 & 1 & 4 & 4 &  &  & 23.1\% \\ 
        & MCore & 64 & 1 & 2 & 4 & 4 &  & 35.3\% \\ 
        & MCore w/ Folding & 64 & 1 & 2 & 4 & 4 & 1 & 39.0\% \\ 
        \midrule
        \multirow{5}{*}{Mixtral-8x22B-G8T8} 
        & FSDP & 128 & 1 & 8 & 1 & & & 2.2\% \\ 
        & FSDP + EP & 128 & 1 & 4 & 8 & & & 9.0\% \\
        & TP + EP + DP & 128 & 1 & 8 & 8 & & & 8.7\% \\
        & MCore & 128 & 1 & 2 & 8 & 8 & & 17.1\% \\
        & MCore w/ Folding & 128 & 1 & 4 & 8 & 8 & 1 & 28.8\% \\
        \midrule
        \multirow{5}{*}{Llama3-8x70B} 
        & FSDP & 256 & 8 & 8 & 1 & & & OOM \\
        & FSDP + EP & 256 & 1 & 8 & 8 & & & 19.6\% \\
        & TP + EP + DP & 256 & 1 & 8 & 8 & & & OOM \\
        & MCore & 256 & 1 & 8 & 4 & 8 & & 38.8\% \\
        & MCore w/ Folding & 256 & 1 & 8 & 8 & 16 & & 41.6\% \\
        \bottomrule
    \end{tabular}
\end{table}
% TODO: clean this table
\begin{longtable}{l l c c}
    \caption{The detailed parallel mapping of scaling experiments for the numbers of GPUs } \label{tab:detailed_GPUs_scaling} \\

    \toprule
    Model & Methods & GPUs & MFU \\ 
    \midrule
    \endfirsthead

    \toprule
    Model & Methods & GPUs & MFU \\ 
    \midrule
    \endhead

    \bottomrule
    \endfoot
    \multirow{16}{*}{Mixtral 8x22B} 
        & \multirow{4}{*}{MCore} & 128 & 49.4\% \\ 
        &                   & 256   & 48.0\% \\ 
        &                   & 512   & 45.5\% \\ 
        &                   & 1024  & 42.3\% \\ 
        \cmidrule(lr){2-4}
        & \multirow{4}{*}{MCore w/ Folding} & 128 & 52.2\% \\ 
        &                   & 256   & 50.7\% \\ 
        &                   & 512   & 48.9\% \\ 
        &                   & 1024  & 44.9\% \\ 
        \cmidrule(lr){2-4}
        & \multirow{4}{*}{FSDP + EP} & 128 & 23.9\% \\ 
        &                   & 256   & 25.5\% \\ 
        &                   & 512   & 24.4\% \\ 
        &                   & 1024  & 23.8\% \\ 
        \cmidrule(lr){2-4}
        & \multirow{4}{*}{TP + EP + DP } & 128 & 40.4\% \\ 
        &                   & 256   & 39.1\% \\ 
        &                   & 512   & 36.0\% \\ 
        &                   & 1024  & 36.2\% \\ 
    \midrule
    
    \multirow{25}{*}{Qwen2 57B-A14B} 
        & \multirow{4}{*}{MCore} & 64 & 36.2\% \\
        &                   & 128   & 36.0\% \\ 
        &                   & 256   & 34.8\% \\ 
        &                   & 512   & 32.5\% \\ 
        &                   & 1024  & 29.8\% \\ 
        \cmidrule(lr){2-4}
        & \multirow{4}{*}{MCore w/ Folding} & 64 & 39.9\% \\
        &                   & 128   & 39.7\% \\ 
        &                   & 256   & 38.1\% \\ 
        &                   & 512   & 36.6\% \\ 
        &                   & 1024  & 33.4\% \\ 
        \cmidrule(lr){2-4}
        & \multirow{4}{*}{FSDP + EP} & 64 & 26.3\% \\
        &                   & 128   & 25.6\% \\ 
        &                   & 256   & 23.4\% \\ 
        &                   & 512   & 22.8\% \\ 
        &                   & 1024  & 21.6\% \\ 
        \cmidrule(lr){2-4}
        & \multirow{4}{*}{TP + EP + DP } & 64 & 22.4 \% \\
        &                   & 128   & 20.9 \% \\ 
        &                   & 256   & 19.8 \% \\ 
        &                   & 512   & 19.7 \% \\ 
        &                   & 1024  & 17.9 \% \\ 

    \midrule

    \multirow{16}{*}{Mixtral 8x22B G8T8} 
        & \multirow{4}{*}{MCore} & 128 & 19.8\% \\ 
        &                   & 256   & 18,4\% \\ 
        &                   & 512   & 16.3\% \\ 
        &                   & 1024  & 13.4\% \\ 
        \cmidrule(lr){2-4}
        & \multirow{4}{*}{MCore w/ Folding} & 128 & 30.0\% \\ 
        &                   & 256   & 29.3\% \\ 
        &                   & 512   & 26.7\% \\ 
        &                   & 1024  & 25.5\% \\ 
        \cmidrule(lr){2-4}
        & \multirow{4}{*}{FSDP + EP} & 128 & 9.0\% \\ 
        &                   & 256   & 8.5\% \\ 
        &                   & 512   & 8.9\% \\ 
        &                   & 1024  & 8.6\% \\ 
        \cmidrule(lr){2-4}
        & \multirow{4}{*}{TP + EP + DP } & 128 & 8.6\% \\ 
        &                   & 256   & 8.5\% \\ 
        &                   & 512   & 8.5\% \\ 
        &                   & 1024  & 8.1\% \\ 

    \midrule
    
    \multirow{9}{*}{Llama3 8x70B} 
        & \multirow{4}{*}{MCore} & 256 & 40.1\% \\ 
        &                   & 512   & 39.5\% \\ 
        &                   & 1024  & 39.1\% \\ 
        \cmidrule(lr){2-4}
        & \multirow{4}{*}{MCore w/ Folding} & 128 & 43.7\% \\ 
        &                   & 512   & 42.7\% \\ 
        &                   & 1024  & 41.5\% \\ 
        \cmidrule(lr){2-4}
        & \multirow{4}{*}{FSDP + EP} & 128 & 18.9\% \\ 
        &                   & 512   & 17.3\% \\ 
        &                   & 1024  & 17.1\% \\ 
\end{longtable}
\begin{table}[h]
    \centering
    \caption{The performance of scaling experiments for the sequence length}
    \label{tab:detailed_context_scaling}
    \begin{tabular}{l l c c c c c c c c c}
        \toprule
        model & methods & \#GPUs & SeqLen & CP & TP & EP & PP & ETP & GBS & MFU \\ 
        \midrule
        \multirow{8}{*}{Mixtral-8x22B} & \multirow{4}{*}{Mcore} & 128 & 16384 & 4 & 2 & 4 & 8 &  & 1024 & 45.30\% \\ 
         & & 256 & 32768 & 8 & 2 & 4 & 8 &  & 512  & 43.20\% \\ 
         & & 512 & 65536 & 16 & 2 & 4 & 8 &  & 256  & 42.60\% \\ 
         & & 1024& 131072& 16 & 4 & 8 & 8 &  & 128  & 38.20\% \\ 
         \cmidrule(lr){2-11}
        & \multirow{4}{*}{Mcore w/ Folding }  & 128   & 16384 & 4 & 2 & 8 & 8 & 1 & 1024 & 47.60\% \\ 
         & & 256 & 32768 & 8 & 2 & 8 & 8 & 1 & 512 & 45.10\% \\ 
         & & 512 & 65536 & 8 & 4 & 8 & 8 & 1 & 256 & 44.50\% \\ 
         & & 1024 & 131072 & 8 & 8 & 8 & 8 & 1 & 128 & 42.90\% \\ 
        \midrule
        \multirow{8}{*}{Qwen2-57B-A14B} & \multirow{4}{*}{Mcore} & 128 & 16384 & 4 & 2 & 4 & 8  &  & 1024 & 45.30\% \\ 
         & &256 &32768 &8 &2 &4 &8 & &512 &43.20\% \\ 
         & &512 &65536 &16 &2 &4 &8 & &256 &42.60\% \\ 
         & &1024 &131072 &16 &4 &8 &8 & &128 &38.20\% \\ 
         \cmidrule(lr){2-11}
        &  \multirow{4}{*}{Mcore w/ Folding} & 128 & 16384 & 4 & 2 & 8 & 8 & 1 & 1024 & 47.60\% \\ 
        & & 256 &32768 &8 &2 &8 &8 &1 &512 &45.10\% \\ 
        & & 512 & 65536 & 8 & 4 & 8 & 8 & 1 & 256 & 44.50\% \\ 
        & & 1024 & 131072 & 8 & 8 & 8 & 8 & 1 & 128 & 42.90\% .\\
        \bottomrule
    \end{tabular}
\end{table}

\end{document}